\documentclass{article} % For LaTeX2e
\usepackage{iclr2027_conference,times}

\usepackage{amsmath,amsfonts,bm}

\def\eqref#1{equation~\ref{#1}}
\def\1{\bm{1}}

\DeclareMathAlphabet{\mathsfit}{\encodingdefault}{\sfdefault}{m}{sl}
\SetMathAlphabet{\mathsfit}{bold}{\encodingdefault}{\sfdefault}{bx}{n}

\usepackage{hyperref}
\usepackage{url}
\usepackage{booktabs}
\usepackage{graphicx}
\usepackage{multirow}
\usepackage{wrapfig}
\usepackage{subcaption}
\usepackage{booktabs}
\usepackage{caption}
\usepackage[table]{xcolor}
\usepackage{array}
\newcolumntype{G}{>{\columncolor{gray!12}}c}
\usepackage[most]{tcolorbox}

\newtcolorbox{observationbox}{
  colback=white,
  colframe=black!75,
  boxrule=0.8pt,
  arc=2mm,
  left=2mm,
  right=2mm,
  top=1mm,
  bottom=1mm,
  before skip=6pt,
  after skip=6pt
}

\title{Balancing Memory Pathways: Analyzing and Improving Memory Utilization in Hybrid LMs}

\author{
\textbf{Hyunji Lee}\textsuperscript{1}\thanks{Correspondence to: \texttt{hyunjil@cs.unc.edu}} \quad
\textbf{Joykirat Singh}\textsuperscript{1} \quad
\textbf{Zaid Khan}\textsuperscript{1} \quad 
\textbf{Justin Chih-Yao Chen}\textsuperscript{1} \quad
\\ 
\textbf{Elias Stengel-Eskin}\textsuperscript{2} \quad
\textbf{Alessandro Sordoni}\textsuperscript{3} \quad
\textbf{Arman Cohan}\textsuperscript{4} \quad
\textbf{Mohit Bansal}\textsuperscript{1} \vspace{5pt}\\
\textsuperscript{1}UNC Chapel Hill \quad
\textsuperscript{2}The University of Texas at Austin \quad
\textsuperscript{3}Mila \quad
\textsuperscript{4}Yale University
}

\newcommand{\jcc}[1]{\textcolor{cyan}{[#1 --JC]}}
\newcommand{\esc}[1]{\textcolor{orange}{[#1 --ESC]}}

\newcommand{\todo}[1]{\textcolor{red}{[#1 --TODO]}}
\newcommand{\zkc}[1]{\textcolor{brown}{[#1 --Zaid]}}
\newcommand{\joy}[1]{\textcolor{violet}{[#1 --Joy]}}
\newcommand{\ac}[1]{\textcolor{magenta}{[#1 --AC]}}

\newif\ifshowcomments
\showcommentstrue % change to \showcommentstrue for the annotated version
\ifshowcomments\else
  \renewcommand{\jcc}[1]{}
  \renewcommand{\esc}[1]{}
  \renewcommand{\todo}[1]{}
  \renewcommand{\zkc}[1]{}
  \renewcommand{\joy}[1]{}
  \renewcommand{\ac}[1]{}
\fi

\iclrfinalcopy % Uncomment for camera-ready version, but NOT for submission.
\begin{document}

\maketitle

\begin{abstract}
Recurrent–attention hybrid language models~(LMs), which interleave attention and recurrent layers, are increasingly used to combine the efficiency of the recurrent layers with the strong performance of attention layers.
Prior work suggests that attention and recurrent layers offer complementary pathways to use past information: attention supports precise memory recall from earlier tokens, while recurrent layers support consolidation of disparate information over long contexts.
However, we observe that simply having access to both pathways does not mean that hybrid LMs are effectively using them. 
When we analyze how much the models rely on each pathway, we find a substantially greater reliance on attention than on the recurrent state.
Standard supervised fine-tuning improves overall performance but does not improve how the two memory pathways are coordinated: the model becomes more reliant on information propagated by attention layers, while its use of information propagated by recurrent layers remains limited.
To encourage better coordination between the two memory pathways, we add an \textit{auxiliary loss} that limits attention's access to earlier context while the recurrent state propagates through the full sequence. 
This objective encourages the model to retain and use information through the recurrent pathway alongside attention. 
It improves overall performance, with particularly strong gains on tasks involving longer contexts or requiring information aggregation, consistent with the strengths of recurrent layers observed in analysis.
Crucially, this imbalance and the benefit of our auxiliary loss generalize: they apply to multiple recurrent-attention LMs in question-answering and agentic tasks, as well as to attention-based LMs that combine different forms of memory. 
Together, our findings show that simply providing multiple memory pathways does not ensure their effective use, and that targeted supervision is needed to better coordinate them.\footnote{Code in \url{https://github.com/amy-hyunji/Balancing-Memory-Pathways}}
\end{abstract}

\section{Introduction}

Recurrent–attention hybrid language models~\citep{qwen35blog, ren2025sambasimplehybridstate, dong2024hymbahybridheadarchitecturesmall, nvidia2025nemotronhfamilyaccurateefficient}, which consist of recurrent and attention layers, leverage efficient recurrent computation and expressive attention memory to allow flexible access to prior context. 
These architectures contain two memory pathways for retaining and using past information: attention and recurrent layers.
Prior work~\citep{lee2025understandingenhancingmambatransformerhybrids, michalak2025attentionneedretrieval, bick2026retrievalawaredistillationtransformerssmhybrids} has analyzed distinct characteristics and strengths of these components: \textit{attention layers} support precise memory recall by directly accessing earlier information, while \textit{recurrent layers} synthesize information distributed across long contexts. 
Extending these findings, we observe that attention plays a larger role when the answer appears directly in the context or can be found from a single piece of evidence, whereas the recurrent state plays a larger role when the answer must be inferred rather than copied or requires combining multiple pieces of evidence across the context.
These complementary strengths motivate an important question: \textbf{How well do hybrid LMs use both memory pathways, and how can their coordination be improved?} 

\begin{figure}[t!]
    \centering
    \begin{minipage}[t]{\linewidth}
    \centering
    \includegraphics[width=\linewidth]{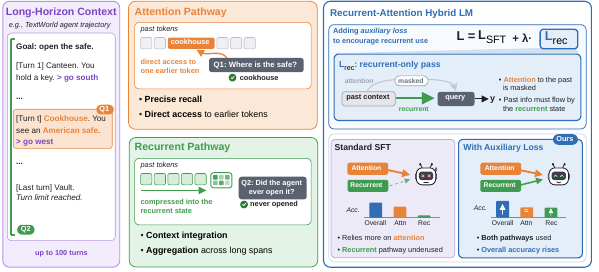}
    \caption{
    Overview of memory pathways in recurrent-attention hybrid LMs and our findings. 
    (\textbf{Middle}) Two memory pathways in recurrent-attention hybrid LMs play different roles: \textit{attention} supports precise memory recall of earlier tokens, while the \textit{recurrent} state supports aggregating information across long contexts. 
    (\textbf{Right}) With Standard SFT~($L_{SFT}$), the model relies mostly on attention and underuses the recurrent pathways. 
    We add an auxiliary loss~($L_{rec}$) that masks attention to past context, allowing the model to access it only through recurrent layers.
    We observe that adding auxiliary loss improves overall performance, with increased use of the recurrent pathway and preserved use of attention.
    }
    \label{fig:main_fig}
    \end{minipage}
\end{figure}

We first examine how existing hybrid LMs use their attention and recurrent memory pathways.
We find that having both pathways does \emph{not} mean that models use both effectively; models retain most of their accuracy even when the recurrent pathway's access to past information is limited 
(30.2\% $\rightarrow$ 20.2\%), whereas accuracy drops substantially when the same restriction is applied to attention layers
(30.2\% $\rightarrow$ 5.1\%).
This imbalance in memory utilization persists even after standard supervised fine-tuning~(SFT) on downstream tasks such as question answering. 
Although the overall performance improves, the gains are mostly from better utilization of attention layers whereas their use of the recurrent pathway remains limited~(Figure~\ref{fig:main_fig}, \textit{Standard SFT} box).

To address this imbalance in memory utilization, we add an \textit{auxiliary loss} that complements standard SFT by encouraging the model to retain and use information through the recurrent pathway.  
In an auxiliary forward pass, we restrict the attention layers' access to past context, requiring the model to access it through recurrent states. 
This simple intervention improves question-answering performance, with average absolute accuracy gains of 3.9\% on Qwen3.5-4B~\citep{qwen35blog} and 5.2\% on Nemotron-H-4B-Instruct~\citep{nvidia2025nemotronhfamilyaccurateefficient} across six datasets.
Adding the auxiliary loss also substantially improves performance when past information is propagated only through recurrent layers, suggesting increased use of recurrent memory.
The gains are particularly large on longer-context datasets~(8.0\%) compared to shorter context (0.2\%), consistent with more effective use of recurrent layers' previously identified strength in processing long contexts.
Reversing the restriction to encourage attention leads to smaller gains, suggesting that the improvement from the auxiliary loss depends on which pathway is encouraged, with greater improvement from strengthening the underutilized recurrent pathway (Figure~\ref{fig:main_fig}, \textit{With Auxiliary Loss} box).

The benefits of auxiliary supervision also extend to agentic tasks, where agents must explore, accumulate observations, and use them to choose subsequent actions.
When averaged across both models, adding recurrent auxiliary supervision improves the success rate by 3.5\% and 9.3\% on Quest and Treasure tasks in TextWorld~\citep{côté2019textworldlearningenvironmenttextbased}, and by 22.0\% and 13.5\% on the PutNextLocal and GoToObjMaze tasks in BabyAI~\citep{chevalierboisvert2019babyaiplatformstudysample}, respectively. 
We observe that the larger gains on BabyAI mirror our question answering results, where the auxiliary loss tends to show stronger improvement on longer contexts; BabyAI trajectories (avg. 134.0 steps) are substantially longer compared to TextWorld~(avg. 42.2 steps).

We further examine how the auxiliary loss affects the use of past information. 
When models are allowed to retry after failures, with their previous attempts stored in memory so that they can revise their answers or actions~\citep{shinn2023reflexionlanguageagentsverbal, zhao2024expelllmagentsexperiential}, models trained with the auxiliary loss recover more successfully, improving the cumulative recovery rate over standard SFT by 16.6\% on question answering and 19.2\% on agentic tasks.
In question answering, the gain is especially large on multi-evidence questions compared to single-evidence questions~(30.3\% vs. 2.8\%). 
We attribute this to the auxiliary loss encouraging greater use of recurrent layers, which we observe to play a larger role in multi-evidence questions.
In the agentic setup, when given multiple suboptimal successful demonstrations, we observe that agents trained with the auxiliary loss more often generate trajectories shorter than any provided demonstrations~(6.0\% on average), with larger gains when the demonstrations are longer. 
Finally, the same imbalance and the strength of the auxiliary loss generalize to attention-based LMs with multiple memories: standard SFT favors one memory pathway over another, and adding auxiliary supervision that encourages the use of the underutilized pathway increases usage of the pathway and improves overall performance in question-answering.
Across architectures and tasks, our findings highlight that \textbf{simply providing multiple memory pathways does \textit{not} lead to balanced, complementary use; models must instead be trained to coordinate these memory pathways.}

\section{Background \& Related Work}
\label{main: background}

In this section, we provide an overview of recurrent-attention hybrid LMs in Section~\ref{background:hybrid_LM}, then review and extend prior findings on the roles of the recurrent and attention memory pathways in Section~\ref{background:memory_roles}. 

\subsection{Recurrent-Attention Hybrid LMs}
\label{background:hybrid_LM}

Recurrent-attention hybrid LMs~\citep{ren2025sambasimplehybridstate, dong2024hymbahybridheadarchitecturesmall, nvidia2025nemotronhfamilyaccurateefficient, zhang2026metismemoryfoundationmodel} 
integrate attention and recurrent layers to combine strong language modeling performance with computational efficiency.
An \textit{attention layer} computes each token's representation by attending to the current and preceding tokens, providing direct access to past information at the cost of quadratic computation in sequence length and KV cache that grows linearly. 
A \textit{linear recurrent layer}, such as Mamba2~\citep{dao2024transformersssmsgeneralizedmodels} or Gated DeltaNet~\citep{yang2025gateddeltanetworksimproving}, instead updates a fixed-size state as new tokens are processed, enabling computation that scales linearly with sequence length and constant memory size. Throughout the paper, we refer to such linear recurrent layers simply as \textit{recurrent layers}.
Many hybrid LMs interleave these layers, combining attention's direct access to earlier tokens with recurrent layers' compact integration of past information~\citep{kimiteam2025kimilinearexpressiveefficient, nvidia2025nemotronhfamilyaccurateefficient}. 
These models often match or outperform fully attention-based models despite using relatively few attention layers, while gaining efficiency from the larger portion of recurrent layers.

\subsection{Roles and Strengths of Attention and Recurrent Memory Pathways}
\label{background:memory_roles}

\paragraph{Findings from Prior Work.}

Prior work has identified distinct roles for attention and recurrent components in hybrid LMs. 
Attention is particularly important for precise retrieval, where removing attention causes severe retrieval failure in the evaluated hybrids~\citep{michalak2025attentionneedretrieval, bick2026retrievalawaredistillationtransformerssmhybrids}. 
Recurrent components contribute substantially to language modeling, where removing the recurrent pathway causes larger degradations in likelihood-based evaluation than removing attention~\citep{lee2025understandingenhancingmambatransformerhybrids, borobia2026componentablationefficienthybrid}.
While these studies characterize the distinct strengths of the two memory pathways, they do not analyze whether hybrid LMs make effective use of both pathways' strengths or how to improve their use.

\begin{figure}[t!]
    \centering
        \begin{minipage}[t]{0.49\linewidth}
        \vspace{0pt}
    \centering
    \includegraphics[width=\linewidth]{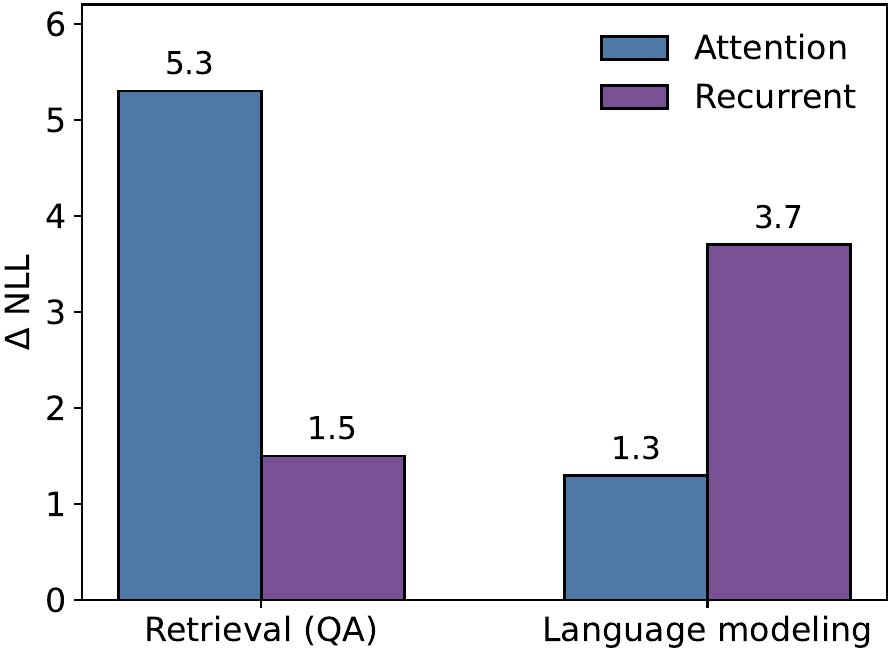}
    \caption{
    Increase in negative log-likelihood (y-axis) for Qwen3.5 when ablating attention or recurrent layers compared to unablated model, where \textit{high} value indicates a \textit{greater contribution of the ablated component}.
    Attention contributes more to retrieval QA, while recurrent layers contribute more to language modeling. 
    }
    \label{fig:hybrid_lm_ret_lm}
    \end{minipage}
    \hfill
\begin{minipage}[t]{0.49\linewidth}
\vspace{0pt}
    \centering
    \includegraphics[width=\linewidth]{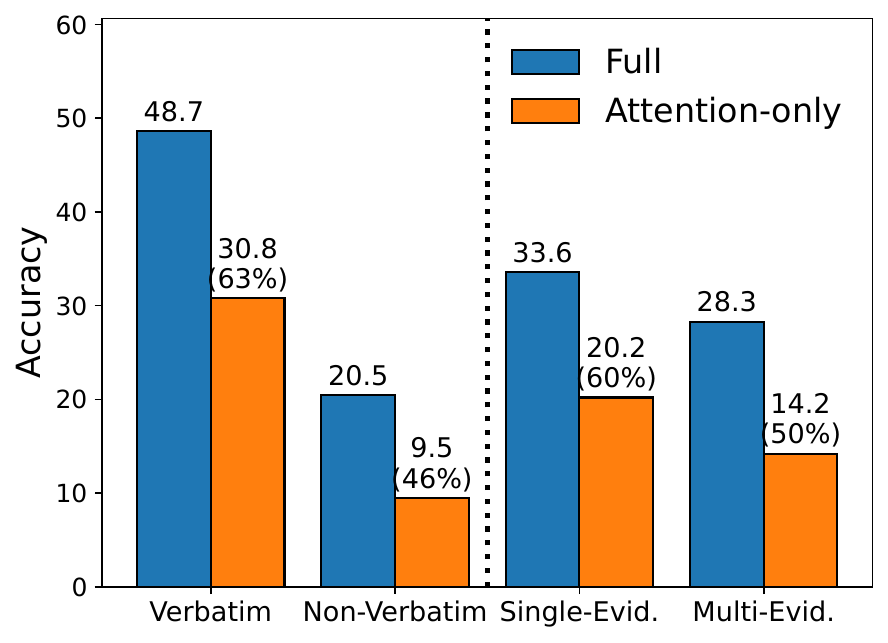}
    \caption{
    Accuracy~(y-axis) of Qwen3.5 under \textit{Full} (unrestricted) and \textit{Attention-only} (past context accessible only through attention) across question types. 
    Percentages show the fraction of Full accuracy retained under Attention-only: high and low retention indicates greater reliance on attention and recurrent state, respectively. 
    }
    \label{fig:recurrent_contribution_analysis}
    \end{minipage}
\end{figure}

\paragraph{Revisiting and Extending Prior Findings.}

We first revisit the distinct roles of attention and recurrent layers identified in the studies reviewed above. 
Following the setup from \citet{lee2025understandingenhancingmambatransformerhybrids}, we use Based NQ~\citep{arora2024justreadtwiceclosing} to evaluate memory recall and PG19~\citep{rae2019compressivetransformerslongrangesequence} to evaluate language modeling, and measure the increase in negative log-likelihood~(NLL) when either attention or recurrent layers are ablated relative to the unablated model using Qwen3.5.
Figure~\ref{fig:hybrid_lm_ret_lm} shows a pattern consistent with prior findings: on retrieval QA, ablating attention increases NLL more than ablating recurrent layers~(5.3 vs. 1.5). 
The pattern reverses for language modeling, where ablating recurrent layers has a larger effect (1.3 vs. 3.7).
This contrast suggests that attention contributes more to precise recall, while recurrent layers contribute more to aggregating information across the context. 

We then extend the analysis to different question types to test whether the two memory pathways have distinct roles or strengths for different information needs using the datasets described in Section~\ref{sec: exp_setup}. 
We consider two distinctions.
First, we classify a question as \textit{verbatim} if its normalized gold answer appears in the source context, and as \textit{non-verbatim} otherwise.
Second, we classify questions as \textit{single-evidence} or \textit{multi-evidence} based on the number of evidence required to answer them, where evidence annotations are available.
We match the number of examples across the two categories in each distinction.
Figure~\ref{fig:recurrent_contribution_analysis} compares pathway use across question types. 
We measure \emph{attention-only retention} as the fraction of full-model accuracy preserved when cross-context information is accessible only through attention layers; higher retention indicates greater reliance on attention and less on recurrent state. 
We observe that retention is higher for verbatim than non-verbatim questions (63\% vs.\ 46\%) and for single-evidence than multi-evidence questions (60\% vs.\ 50\%).
This aligns with the findings above: attention layers play a larger role in recovering precise details, but contribute less when information must be abstracted or integrated across the context. 
Further details are in Appendix~\ref{app:role}.

\section{Experimental Setup}
\label{sec: exp_setup}
\paragraph{Models.}
We experiment over two recurrent-attention hybrid LMs: \texttt{Qwen3.5-4B}~\citep{qwen35blog} and \texttt{Nemotron-H-4B-Instruct}~\citep{nvidia2025nemotronhfamilyaccurateefficient}. 
\texttt{Qwen3.5-4B} contains 24 \textit{Gated DeltaNet}~\citep{yang2025gateddeltanetworksimproving} and 8 full-attention layers. 
\texttt{Nemotron-H-4B-Instruct} consists of 24 \texttt{Mamba2}~\citep{dao2024transformersssmsgeneralizedmodels} and 4 full-attention layers.

\paragraph{Training.}
We train all models using LoRA~\citep{hu2021loralowrankadaptationlarge} on all layers with rank 16, scaling factor 32, and dropout 0.05.
We use AdamW with a learning rate of $1e^{-4}$, a cosine learning rate schedule, a warmup period covering 3\% of training steps, and an effective batch size of 16.
Since objectives with an auxiliary loss require additional computation, we halve the number of optimization steps to match the total training compute with standard SFT. 
Unless otherwise specified, we use the same optimization setting across all compared variants.

\paragraph{Task \& Evaluations.}
We experiment on long-context question answering and long-horizon agentic tasks, which are two common settings for studying memory in which the model needs to retain and use information from earlier contexts or interactions for strong performance~\citep{packer2024memgptllmsoperatingsystems,zhao2024expelllmagentsexperiential,shinn2023reflexionlanguageagentsverbal}.
For \textbf{Question Answering}, we pool training examples from four datasets so that the model is trained over diverse context lengths and reasoning types: DROP~\citep{dua2019dropreadingcomprehensionbenchmark}, HotpotQA~\citep{yang2018hotpotqadatasetdiverseexplainable}, Qasper~\citep{dasigi2021datasetinformationseekingquestionsanswers}, NarrativeQA~\citep{kočiský2017narrativeqareadingcomprehensionchallenge}.
This yields 12k training and 2k validation instances.
We evaluate each dataset separately, and additionally evaluate on MINTEval~\citep{lee2026mintevalevaluatingmemorymultitarget} and MuSR~\citep{sprague2024musrtestinglimitschainofthought}, leading to a total of 8.2k test examples with an average context length of 84k tokens.
For \textbf{Agentic Tasks}, we experiment over TextWorld~\citep{côté2019textworldlearningenvironmenttextbased} and BabyAI~\citep{chevalierboisvert2019babyaiplatformstudysample}, where agents must explore the environment, accumulate information through sequences of observations and actions, and use it to make subsequent decisions. The environments differ in their observation spaces, action spaces, and in the structure of their trajectories.
In TextWorld, we train on Quest and evaluate on both Quest and Treasure.
In BabyAI, we train on MiniBossLevel and evaluate on PutNextLocal and GoToObjMaze. 
Further details are in Appendix~\ref{app:exp_setup}.

\section{Standard SFT Favors Attention over Recurrence}
\label{main: sft}

\begin{observationbox}
\textbf{Key Observation:} 
Standard SFT increases recurrent-attention hybrid LMs’ use of the attention pathway, but yields little improvement in their use of the recurrent pathway.
\end{observationbox}

In Section~\ref{background:memory_roles}, we observed complementary strengths and roles of the two pathways: attention supports precise recall, while recurrent layers integrate information across the context.
Thus, in this section, we examine whether standard supervised fine-tuning~(SFT) leads the model to use both pathways in ways that leverage these complementary strengths.
We first define \textit{attention-only} and \textit{recurrent-only} conditions to analyze and assess the model's use of each memory pathway~(Section~\ref{sft:disentangle}).
We then evaluate how overall performance and pathway utilization change after SFT on question answering and agentic tasks~(Section~\ref{sft:results}).

\subsection{Disentangling two memory pathways}
\label{sft:disentangle}
To analyze how past information is carried through attention and recurrence, we divide each input into a \text{past} portion and a \text{current} portion, and selectively restrict which memory mechanism can transfer information from the \text{past} to the \text{current} input.
For example, in question answering, the past portion corresponds to the context and the current portion to the question.
We use \textbf{full} to denote the unmodified condition, in which both attention and recurrent layers can propagate information from the past to the current portion.
We construct two complementary interventions, \textbf{recurrent-only} and \textbf{attention-only}; these indicate \textbf{which mechanism can carry information from the past to the current portion}.
In the recurrent-only setting, attention cannot access the past portion, so past information reaches the current portion only through the recurrent state; the attention-only setting does the opposite. In both, all layers remain active within each portion.

Specifically, in the \textit{recurrent-only} condition, we block attention across segment boundaries while allowing the recurrent state to propagate normally:
\[
M_{t,s}
=
\begin{cases}
0, & s \leq t \text{ and } \sigma(s)=\sigma(t),\\
-\infty, & \text{otherwise}.
\end{cases}
\]
$M_{t,s}$ is added to the attention score between query position $t$ and key position $s$ before normalization, and $\sigma(t)$ denotes the segment containing token $t$.
The mask therefore retains standard causal attention within each segment, while blocking attention to all tokens in preceding segments.

Conversely, in the \textit{attention-only} condition, we retain the standard causal attention over the preceding context, while restricting the recurrent state from propagating across segment boundaries ($\sigma(\cdot)$). For $t>1$, we modify the recurrent update as:
\[
S_t
=
\mathbb{1}\!\left[\sigma(t)=\sigma(t-1)\right]
\mathcal{A}_t(S_{t-1})
+
w_t,
\qquad
S_0 = 0,
\]
where $S_t$ denotes the recurrent state after processing token $t$, $\mathcal{A}_t$ denotes the recurrent transition applied to the previous state, $w_t$ denotes the contribution of the current token, and $\mathbb{1}[\cdot]$ denotes an indicator function.
At a segment boundary ($\sigma(t)\neq\sigma(t-1)$), the indicator is zero, so the previous recurrent state does not contribute to the new segment and within a segment, the recurrent update is unchanged.
Thus, information from preceding segments remains accessible only through attention.

\begin{figure}[t!]
\vspace{-2em}
    \centering
    \begin{minipage}[t]{0.49\linewidth}
    \centering
    \includegraphics[width=\linewidth]{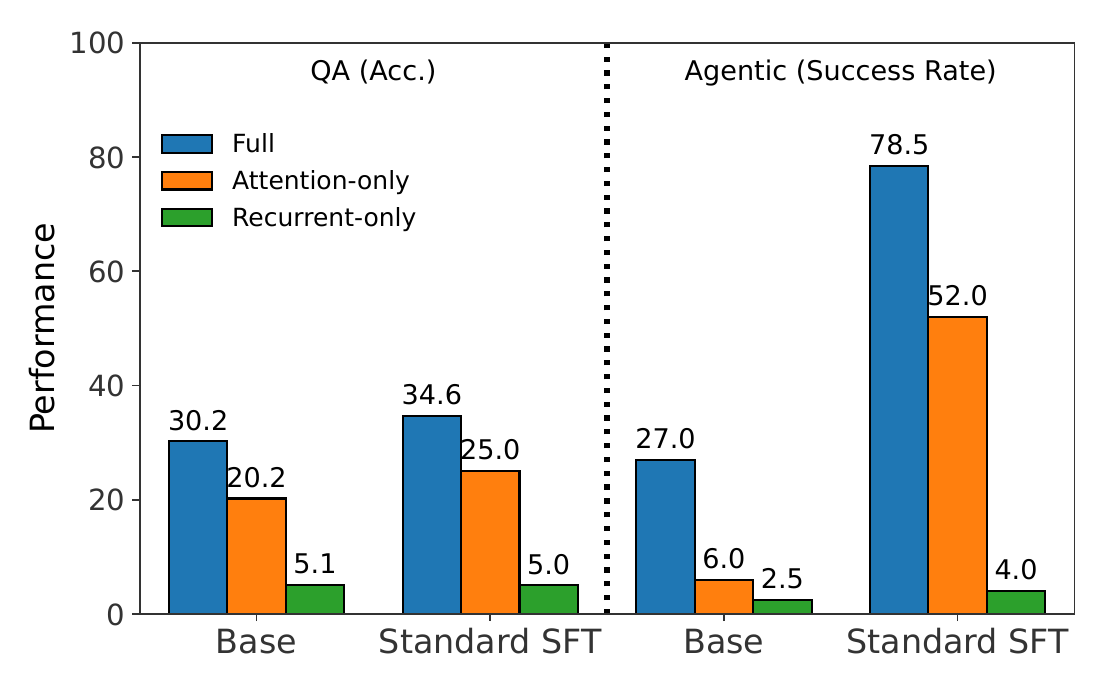}
    \caption{
    Performance of Qwen3.5-4B before~(\textit{Base}) and after standard SFT~(\textit{Standard SFT}) on question answering and agentic tasks.
    \textit{Standard SFT} substantially improves \textit{Full} and \textit{Attention-only} performance but little improvement in \textit{Recurrent-only} performance.
    }
    \label{fig:qa_underutilization}
    \end{minipage}
    \hfill
\begin{minipage}[t]{0.49\linewidth}
    \centering
    \includegraphics[width=\linewidth]{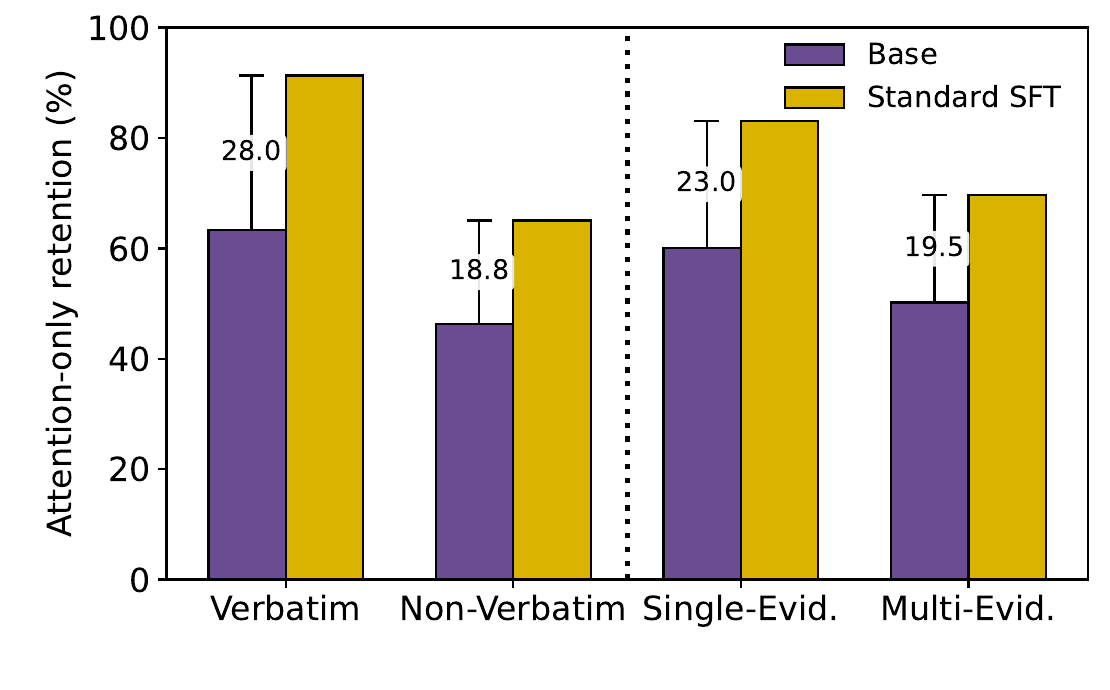}
    \caption{
    Attention-only retention (the fraction of \textit{Full} accuracy retained when past context is accessible only through attention) before (\textit{Base}) and after SFT~(\textit{Standard SFT}). 
    SFT increases attention-only retention across all question types, especially verbatim and single-evidence. 
    }
    \label{fig:sft_recurrent_contribution_analysis}
    \end{minipage}
\end{figure}

\subsection{Memory Pathway Utilization after SFT}
\label{sft:results}
We evaluate both the original and SFT-trained models under the full, attention-only, and recurrent-only conditions defined in Section~\ref{sft:disentangle}.
As shown in Figure~\ref{fig:qa_underutilization}, the SFT-trained model~(\textit{Standard SFT}) achieves 34.6\% QA accuracy on Qwen3.5 under the full condition, compared to 30.2\% for the original model~(\textit{Base}).
Attention-only accuracy also increases from 20.2\% to 25.0\%, whereas recurrent-only accuracy remains nearly unchanged~(5.1\% to 5.0\%).
We observe a similar pattern on the agentic Quest task: full success rate rises from 27.0\% to 78.5\%, attention-only also rises from 6.0\% to 52.0\%, while recurrent-only improves only slightly, from 2.5\% to 4.0\%.
These results indicate that the model retains a large portion of its full performance when recurrent layers are restricted to local information (attention-only), but drops sharply when attention is restricted to local information (recurrent-only). 
This gap widens after SFT.

This shift toward the attention pathway also appears across the question types discussed in Section~\ref{background:memory_roles}. 
We observe that attention-only retention, the fraction of full-model accuracy preserved under the attention-only condition, increases for all question types~(Figure~\ref{fig:sft_recurrent_contribution_analysis}). 
The gain is larger for question types that already relied more on attention: verbatim and single-evidence questions show an average increase of 25.5\% compared to 19.2\% for non-verbatim and multi-evidence questions.

Together, these results show that standard SFT improves overall performance but does not lead to balanced use of the two memory pathways: the model becomes more reliant on attention, while its use of recurrent pathway changes little.
Given the distinct strengths of these pathways observed in Section~\ref{background:memory_roles}, we next investigate whether encouraging more effective use of both pathways can improve memory coordination and further enhance overall performance. 

\section{Encouraging Recurrent Memory Use Improves Performance}
\label{hybrid_model: auxiliary_loss}

\begin{observationbox}
\textbf{Key Observation:} 
Adding an auxiliary loss that encourages the use of recurrent pathway increases its utilization and improves overall performance while maintaining utilization of the attention pathway.
\end{observationbox}

In Section~\ref{background:memory_roles}, we observe that attention and recurrent layers play complementary roles, each carrying important information for different types of questions. 
However, Section~\ref{sft:results} shows that standard SFT does not effectively utilize both pathways, instead increasing the model's reliance on attention while leaving the recurrent pathway underutilized. 
Thus, in this section, we examine whether adding auxiliary loss that encourages the use of recurrent pathway~(Section~\ref{auxiliary_loss}) leads to better memory coordination and stronger overall performance~(Section~\ref{aux_loss:results}).

\subsection{Recurrent Auxiliary Supervision}
\label{auxiliary_loss}
We augment standard SFT with an auxiliary loss that uses the same supervised next-token prediction objective but masks attention to the preceding context, a mechanism commonly used to control information flow in LMs~\citep{mu2024learningcompresspromptsgist, dong2019unifiedlanguagemodelpretraining}.
Unlike attention, recurrent layers remain unrestricted: they process the entire sequence and propagate their state from the preceding context. 
Thus, during the auxiliary forward pass, information from the preceding context can influence the answer only through recurrent memory, encouraging the model to retain and use information throughout the recurrent pathway.
We jointly optimize the standard SFT loss~($\mathcal{L}_{\mathrm{SFT}}(\theta)$), computed without these restrictions, and the auxiliary loss~($\mathcal{L}_{\mathrm{rec}}(\theta)$): 
\[
\mathcal{L}(\theta)
=
\mathcal{L}_{\mathrm{SFT}}(\theta)
+
\lambda\,\mathcal{L}_{\mathrm{rec}}(\theta),
\]
where $\theta$ denotes model parameters and $\lambda$ controls the strength of the auxiliary supervision.\footnote{Ablation of different $\lambda$ in Appendix~\ref{app_hybrid_model:lambda}}

\subsection{Results}
\label{aux_loss:results}

\begin{table}[t]
\vspace{-2em}
\centering
\small
\caption{
QA accuracy across six datasets.
First two rows compare standard SFT with our objective; the remaining rows ablate the training objective.
Avg. is micro-average over all test examples.
}
\begin{tabular}{ll|cccccc|G}
\toprule
Model & Training Obj. & DROP & MuSR & HQA & Qasper & NQA & MINT & Avg. \\
\midrule
\multirow{6}{*}{\texttt{Qwen3.5}}
& $\mathcal{L}_{\mathrm{SFT}}$
& 48.6 & 45.6 & 43.1 & 30.1 & 24.4 & 21.1 & 34.6 \\
& $\mathcal{L}_{\mathrm{SFT}}+\mathcal{L}_{\mathrm{rec}}$ (Ours)
& 46.9 & 44.8 & 44.0 & \textbf{35.9} & \textbf{33.7} & \textbf{28.9} & \textbf{38.5} \\
\cmidrule{2-9}
& $\mathcal{L}_{\mathrm{rec}}$
& 35.0 & 36.5 & 38.0 & 32.1 & 20.1 & 23.6 & 30.4 \\
& $\mathcal{L}_{\mathrm{attn}}$
& \textbf{56.9} & 48.1 & 35.7 & 31.3 & 15.6 & 10.3 & 31.6 \\
& $\mathcal{L}_{\mathrm{SFT}}+\mathcal{L}_{\mathrm{attn}}$
& 54.0 & \textbf{49.1} & \textbf{45.7} & 30.3 & 23.7 & 17.1 & 35.6 \\
\midrule
\multirow{6}{*}{\texttt{Nemotron-H}}
& $\mathcal{L}_{\mathrm{SFT}}$
& 42.4 & 40.9 & 38.7 & 28.1 & 21.9 & 18.8 & 31.0 \\
& $\mathcal{L}_{\mathrm{SFT}}+\mathcal{L}_{\mathrm{rec}}$ (Ours)
& 44.5 & 38.1 & \text{42.1} & \textbf{34.1} & \textbf{31.7} & \textbf{27.8} & \textbf{36.2} \\
\cmidrule{2-9}
& $\mathcal{L}_{\mathrm{rec}}$
& 31.6 & 37.6 & \textbf{42.8} & 28.0 & 24.1 & 18.4 & 29.8 \\
& $\mathcal{L}_{\mathrm{attn}}$
& 44.7 & \textbf{47.6} & 35.8 & 29.4 & 24.0 & 14.7 & 31.4 \\
& $\mathcal{L}_{\mathrm{SFT}}+\mathcal{L}_{\mathrm{attn}}$
& \textbf{50.7} & 47.1 & 40.1 & 29.0 & 26.1 & 18.0 & 34.1 \\
\bottomrule
\end{tabular}
\label{tab:qa_hybrid_aux}
\end{table}

\begin{figure}[t!]
\vspace{-2em}
    \centering
    \begin{minipage}[t]{0.99\linewidth}
    \centering
    \includegraphics[width=\linewidth]{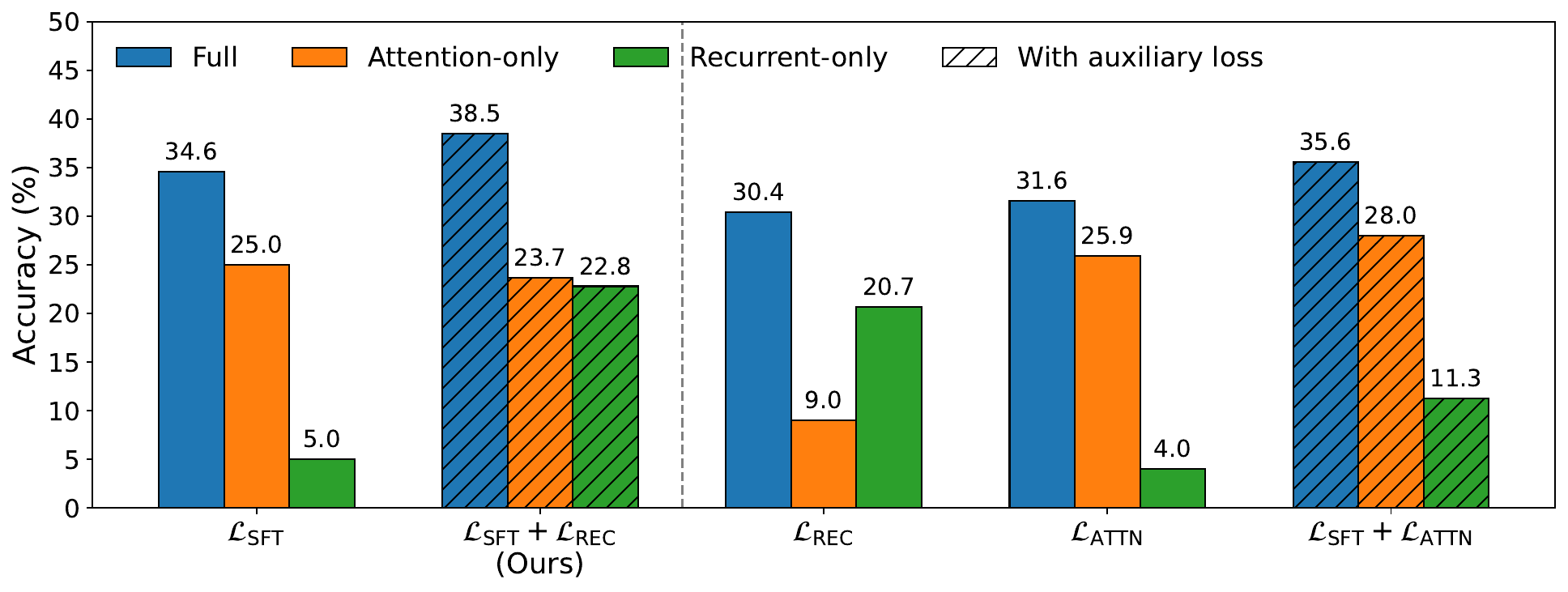}
    \caption{
    QA accuracy of Qwen3.5-4B across various training objectives~(details of objectives in Section~\ref{aux_loss:results}).
    Hatched bars indicate that it adds an auxiliary loss.
    \textit{SFT+REC} increases Recurrent-only accuracy while largely preserving Attention-only accuracy and best Full accuracy. 
    }
    \label{fig:qwen3.5_qa_ablation}
    \end{minipage}
\end{figure}

\paragraph{Question Answering.}
Table~\ref{tab:qa_hybrid_aux} shows accuracy across six question-answering datasets.
When comparing standard SFT~($L_{SFT}$) with when adding auxiliary recurrent supervision~($L_{SFT}$+$L_{rec}$), we observe that adding recurrent auxiliary loss improves the average accuracy from 34.6\% to 38.5\% in Qwen3.5 and from 31.0\% to 36.2\% in Nemotron-H. 
The gains are particularly large on Qasper, NarrativeQA, and MINTEval, suggesting that the auxiliary loss is especially helpful for answering questions from longer contexts.
Figure~\ref{fig:qwen3.5_qa_ablation} further shows that adding recurrent auxiliary supervision substantially improves recurrent-only accuracy (5.0\% to 22.8\%) while largely preserving attention-only accuracy (25.0\% to 23.7\%), suggesting more effective use of recurrent state.
This improved utilization, together with the recurrent pathway's strength in aggregating information across long contexts observed in Section~\ref{background:memory_roles}, suggests that more effective use of recurrent state contributes to the overall performance gains.

To understand which part of the objective drives these gains, we conduct analysis with several training variants.
For these ablations, we consider $L_{attn}$, which is the counterpart of $L_{rec}$: it blocks recurrent-state propagation from the preceding context while leaving attention unrestricted.
We compare training with auxiliary losses alone~($L_{rec}$ or $L_{attn}$), and adding $L_{attn}$ to standard SFT~($L_{SFT}+L_{attn}$).
We observe that $L_{SFT}+L_{attn}$ yields a smaller gain than $L_{SFT}+L_{rec}$ for both models, with average absolute improvement of 2.1\% and 4.6\%, respectively.
Figure~\ref{fig:qwen3.5_qa_ablation} further shows that adding $L_{attn}$ increases attention-only accuracy from 25.0\% to 28.0\%, while recurrent-only accuracy remains much lower at 11.3\%, indicating continued imbalance between the two memory pathways. 
These results suggest that strengthening the underutilized recurrent pathway is more effective than further encouraging the use of attention.
Finally, training with only the auxiliary loss~($L_{attn}$ or $L_{rec}$) yields lower average accuracy than standard SFT~($L_{SFT}$) when averaged over both models, by 1.3\% and 2.7\%, respectively. This suggests that restricted losses are most effective as auxiliary objectives alongside unrestricted SFT.

\begin{table}[t]
\centering
\small
\caption{
Performance on TextWorld and BabyAI under different training objectives.
For each model, the first two rows compare standard SFT with our objective; the remaining rows ablate the training objective. 
Best in \textbf{bold}.
}
\begin{tabular}{ll|cc|cc}
\toprule
\multirow{2}{*}{Model} & \multirow{2}{*}{Training Obj.} 
& \multicolumn{2}{c|}{TextWorld}
& \multicolumn{2}{c}{BabyAI} \\
\cmidrule{3-4} \cmidrule{5-6}
& & Quest & Treasure & PutNextLocal & GoToObjMaze \\
\midrule
\multirow{5}{*}{\texttt{Qwen3.5}}
& $\mathcal{L}_{\mathrm{SFT}}$
& 78.5 & 44.0 & 46.0 & 63.0\\
& $\mathcal{L}_{\mathrm{SFT}}+\mathcal{L}_{\mathrm{rec}}$ (Ours)
& \textbf{80.5} & \textbf{56.0}& \textbf{73.0} & \textbf{83.0}  \\
\cmidrule{2-6}
& $\mathcal{L}_{\mathrm{rec}}$
& 75.5 & 45.0& 13.0& 37.0 \\
& $\mathcal{L}_{\mathrm{attn}}$
& 76.5 & 44.0 & 31.0 & 50.0 \\
& $\mathcal{L}_{\mathrm{SFT}}+\mathcal{L}_{\mathrm{attn}}$
& 77.0 & 49.0 & 47.0 & 70.0\\
\midrule
\multirow{5}{*}{\texttt{Nemotron-H}}
& $\mathcal{L}_{\mathrm{SFT}}$
& 70.5 & 40.5 & 30.0& 77.0 \\
& $\mathcal{L}_{\mathrm{SFT}}+\mathcal{L}_{\mathrm{rec}}$ (Ours)
& \textbf{75.5} & \textbf{47.0}  & \textbf{47.0} & \textbf{84.0}\\
\cmidrule{2-6}
& $\mathcal{L}_{\mathrm{rec}}$
& 68.5 & 40.0 & 17.0 & 20.0   \\
& $\mathcal{L}_{\mathrm{attn}}$
& 71.5 & 42.0 & 37.0 & 54.0 \\
& $\mathcal{L}_{\mathrm{SFT}}+\mathcal{L}_{\mathrm{attn}}$
& 72.0 & 44.5 & 43.0 & 80.0\\
\bottomrule
\end{tabular}
\vspace{-0.5em}
\label{tab:agentic_hybrid_aux}
\end{table}

\paragraph{Agentic Tasks.}
Table~\ref{tab:agentic_hybrid_aux} shows performance on two agentic tasks: TextWorld and BabyAI.
We observe similar trend as we observed from question answering task: when adding recurrent auxiliary supervision to standard SFT~($\mathcal{L}_{\mathrm{SFT}}+\mathcal{L}_{\mathrm{rec}}$), it improves success rate over standard SFT~($\mathcal{L}_{\mathrm{SFT}}$) by a large gain: when averaged across Qwen3.5 and Nemotron-H, the absolute gains are 3.5\% and 9.3\% in TextWorld Quest and Treasure, respectively, and 22.0\% and 13.5\% in BabyAI PutNextLocal and GoToObjMaze, respectively. 
Similarly, $\mathcal{L}_{\text{SFT}}+\mathcal{L}_{\text{attn}}$ yields a smaller gain than $\mathcal{L}_{\text{SFT}}+\mathcal{L}_{\text{rec}}$ across all four games, with average absolute improvements of 4.1\% and 12.1\%, respectively. 
The larger gains on BabyAI are consistent with our question-answering results, where recurrent auxiliary supervision is particularly beneficial on longer-context datasets. 
BabyAI trajectories are generally longer than those in TextWorld, averaging 134.0 versus 42.2 steps across all trajectories and 40.2 versus 13.8 steps among successful trajectories.

\section{Auxiliary Supervision Generalizes}
\label{generalizes}
In this section, we examine whether encouraging the use of the recurrent pathway also improves the model's ability to learn from past failure and successful trajectories~\citep{zhao2024expelllmagentsexperiential, shinn2023reflexionlanguageagentsverbal}. We then examine whether our observations generalize to attention-based LMs with multiple memories.

\begin{figure}[t]
\vspace{-2.5em}
    \centering
    \includegraphics[width=0.9\linewidth]{
        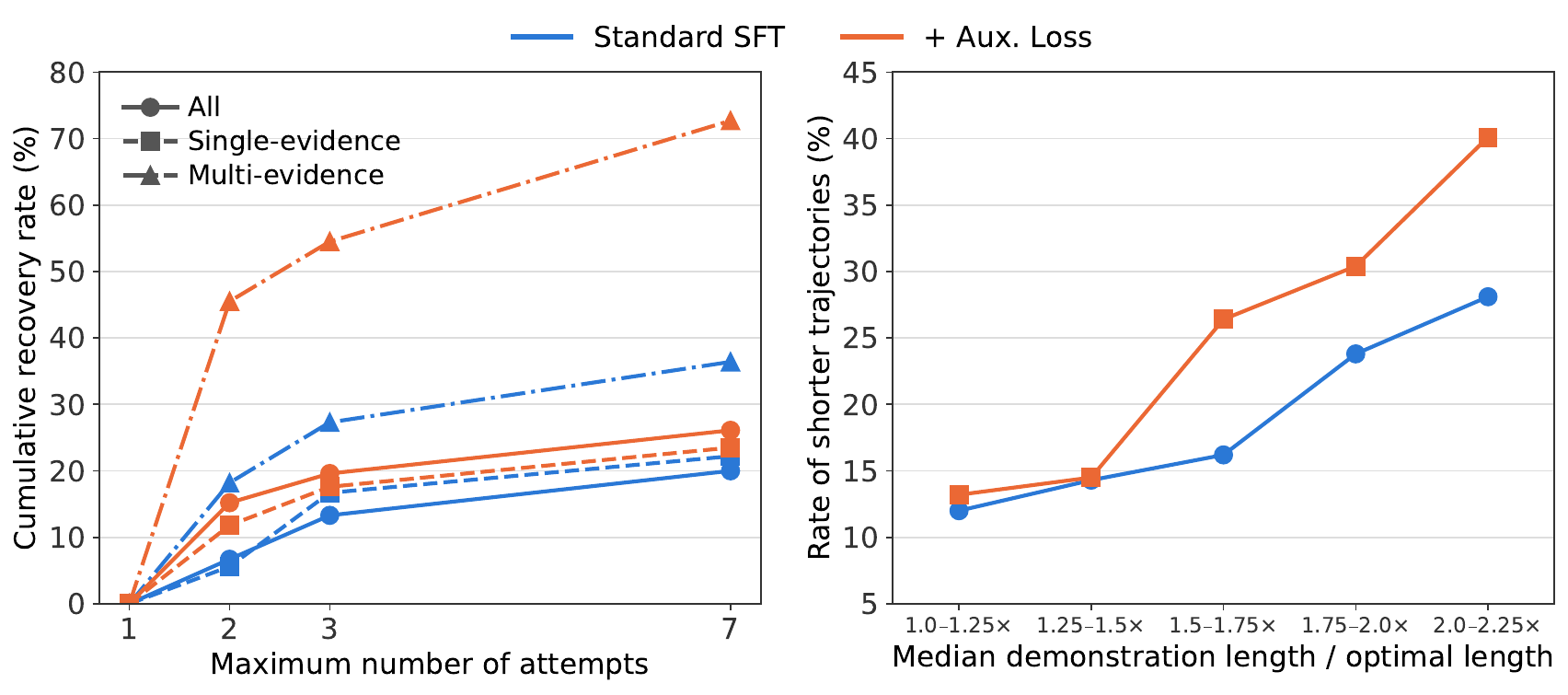
    }
    \caption{
    Auxiliary supervision improves learning from past trajectories.
    \textbf{Left:} Cumulative recovery rate on QA as the maximum
    number of attempts increases, with previous failed attempts
    provided in context. The auxiliary loss improves recovery,
    especially on multi-evidence questions.
    \textbf{Right:} Rate of generating trajectories shorter than
    all provided successful demonstrations, grouped by median
    demonstration length relative to the optimal path.
    The auxiliary loss yields shorter trajectories more often,
    especially with longer demonstrations.
    }
    \label{fig:trajectory_memory}
    \vspace{-0.8em}
\end{figure}

\paragraph{Learning from Failure Trajectories.}
We analyze whether models can recover from their own errors when given additional attempts, with previous failed attempts provided in context~\citep{shinn2023reflexionlanguageagentsverbal,gou2024criticlargelanguagemodels}.
For question answering, we evaluate on MINTEval, which includes long contexts and both single-evidence and multi-evidence questions about the same context.
Figure~\ref{fig:trajectory_memory} (left) shows the cumulative recovery rate (y-axis) as the maximum number of attempts~(x-axis) increases.
Across all questions (solid lines), recovery rates increase with additional attempts under both standard SFT and SFT with auxiliary loss. 
Adding the auxiliary loss consistently improves recovery, with an average absolute gain of 16.6\% across attempt budgets.
The gains are particularly large on multi-evidence questions: averaged across attempt budgets, recovery improves from 27.3\% to 57.6\%, compared to 14.8\% to 17.6\% on single-evidence questions.
For agentic tasks, we apply the same retry setup to TextWorld, providing previous failed trajectories in context. 
The auxiliary loss improves cumulative recovery by an average of 19.2\%~(Figure~\ref{fig:agentic_learning_from_failure} in Appendix). 
Recovery continues to improve with additional attempts under SFT with the auxiliary loss, whereas standard SFT plateaus.
These results suggest that the auxiliary loss helps models more effectively use failed attempts stored in memory to recover from errors.

\paragraph{Learning from Successful Trajectories.}
\label{analysis:agentic}

We examine whether agents can use successful but suboptimal trajectoreis to develop more efficient strategies. 
We provide multiple successful demonstrations of varying lenghts, which may contain unnecessary steps, and measure how often the agent generates a successful trajectory shorter than the shortest demonstration.
As shown in Figure~\ref{fig:trajectory_memory} (right), adding the auxiliary loss improves this rate by an average of 6.0\%, with larger gains as demonstrations length increases relative to the optimal path. 
These results suggest that the auxiliary loss helps agents more effectively use successful demonstrations stored in memory to develop more efficient strategies.

\paragraph{Generalizes to Attention-Based LMs.}

\begin{wrapfigure}{r}{0.45\linewidth}
    \centering
    \vspace{-25pt}
    \includegraphics[width=\linewidth]{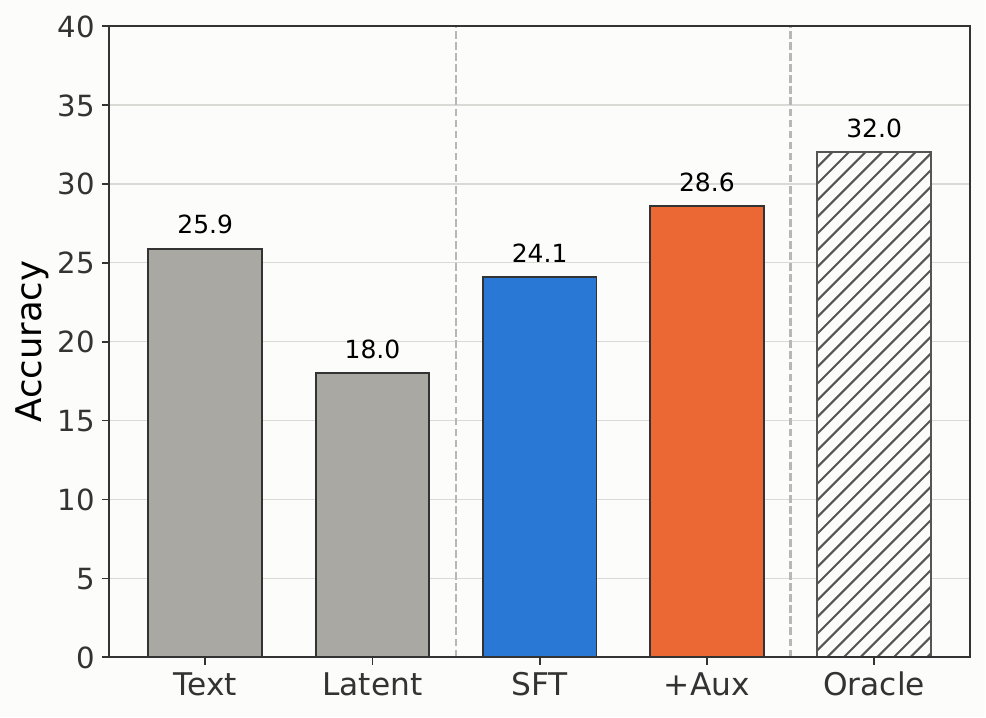}
    \caption{
    QA accuracy of attention-based models given a single memory type (\textit{Text}, \textit{Latent}), both (\textit{SFT}, \textit{+Aux}), or per-example \textit{Oracle} upper bound.
    }
    \label{fig:hybrid_memory_all}
\end{wrapfigure}

We find that both the imbalance in memory use and the benefits of auxiliary supervision extend to attention-based LMs with multiple memory types~(textual and latent memory). 
As shown in Figure~\ref{fig:hybrid_memory_all}, standard SFT~(\textit{SFT}) with both memories achieves 24.1\% accuracy, far below the \textit{Oracle}, which selects the better of the two single-memory predictions for each example~(32.0\%) and even below the text-only model~(25.9\%).
This suggests that access to both memories does not ensure effective use of their complementary strengths. 
Even when trained jointly with standard SFT, the model remains biased toward using textual memory. 
Adding an auxiliary loss that encourages the use of both memory pathways improves accuracy to 28.6\%.
Further analysis shows that this improvement is accompanied by increased use of latent memory while largely preserving use of textual memory~(Appendix~\ref{app:attention_based}).

\section{Conclusion}
In this paper, we show that recurrent-attention hybrid LMs tend to rely predominantly on the attention pathway despite the complementary strengths of their recurrent layers, and that standard SFT further increases attention use while yielding little improvement in recurrent memory use. 
We show that adding a simple auxiliary loss that encourages use of the recurrent pathway improves utilization of both memory pathways and overall performance. 
These findings generalize across multiple recurrent-attention hybrid LMs, across question answering and agentic tasks, and to attention-based LMs that combine different forms of memory.

\section*{Acknowledgment}
This work was supported by NSF-AI Engage Institute DRL-2112635, NSF-CAREER Award 1846185, Capital One Research Award, and NDSEG PhD Fellowship. The views contained in this article are those of the authors and not of the funding agency.

\bibliography{iclr2027_conference}

@misc{packer2024memgptllmsoperatingsystems,
      title={MemGPT: Towards LLMs as Operating Systems}, 
      author={Charles Packer and Sarah Wooders and Kevin Lin and Vivian Fang and Shishir G. Patil and Ion Stoica and Joseph E. Gonzalez},
      year={2024},
      eprint={2310.08560},
      archivePrefix={arXiv},
      primaryClass={cs.AI},
      url={https://arxiv.org/abs/2310.08560}, }

@misc{rae2019compressivetransformerslongrangesequence,
      title={Compressive Transformers for Long-Range Sequence Modelling}, 
      author={Jack W. Rae and Anna Potapenko and Siddhant M. Jayakumar and Timothy P. Lillicrap},
      year={2019},
      eprint={1911.05507},
      archivePrefix={arXiv},
      primaryClass={cs.LG},
      url={https://arxiv.org/abs/1911.05507}, 
}

@misc{gou2024criticlargelanguagemodels,
      title={CRITIC: Large Language Models Can Self-Correct with Tool-Interactive Critiquing}, 
      author={Zhibin Gou and Zhihong Shao and Yeyun Gong and Yelong Shen and Yujiu Yang and Nan Duan and Weizhu Chen},
      year={2024},
      eprint={2305.11738},
      archivePrefix={arXiv},
      primaryClass={cs.CL},
      url={https://arxiv.org/abs/2305.11738}, 
}

@misc{arora2024justreadtwiceclosing,
      title={Just read twice: closing the recall gap for recurrent language models}, 
      author={Simran Arora and Aman Timalsina and Aaryan Singhal and Benjamin Spector and Sabri Eyuboglu and Xinyi Zhao and Ashish Rao and Atri Rudra and Christopher Ré},
      year={2024},
      eprint={2407.05483},
      archivePrefix={arXiv},
      primaryClass={cs.CL},
      url={https://arxiv.org/abs/2407.05483}, 
}

@misc{wang2025memalphalearningmemoryconstruction,
      title={Mem-$\alpha$: Learning Memory Construction via Reinforcement Learning}, 
      author={Yu Wang and Ryuichi Takanobu and Zhiqi Liang and Yuzhen Mao and Yuanzhe Hu and Julian McAuley and Xiaojian Wu},
      year={2025},
      eprint={2509.25911},
      archivePrefix={arXiv},
      primaryClass={cs.CL},
      url={https://arxiv.org/abs/2509.25911}, 
}

@misc{dong2019unifiedlanguagemodelpretraining,
      title={Unified Language Model Pre-training for Natural Language Understanding and Generation}, 
      author={Li Dong and Nan Yang and Wenhui Wang and Furu Wei and Xiaodong Liu and Yu Wang and Jianfeng Gao and Ming Zhou and Hsiao-Wuen Hon},
      year={2019},
      eprint={1905.03197},
      archivePrefix={arXiv},
      primaryClass={cs.CL},
      url={https://arxiv.org/abs/1905.03197}, 
}

@misc{qwen35blog,
    title = {Qwen3.5: Towards Native Multimodal Agents},
    url = {https://qwen.ai/blog?id=qwen3.5},
    author = {{Qwen Team}},
    month = {February},
    year = {2026}
}

@misc{yu2026agenticmemorylearningunified,
      title={Agentic Memory: Learning Unified Long-Term and Short-Term Memory Management for Large Language Model Agents}, 
      author={Yi Yu and Liuyi Yao and Yuexiang Xie and Qingquan Tan and Jiaqi Feng and Yaliang Li and Libing Wu},
      year={2026},
      eprint={2601.01885},
      archivePrefix={arXiv},
      primaryClass={cs.CL},
      url={https://arxiv.org/abs/2601.01885}, 
}

@misc{shinn2023reflexionlanguageagentsverbal,
      title={Reflexion: Language Agents with Verbal Reinforcement Learning}, 
      author={Noah Shinn and Federico Cassano and Edward Berman and Ashwin Gopinath and Karthik Narasimhan and Shunyu Yao},
      year={2023},
      eprint={2303.11366},
      archivePrefix={arXiv},
      primaryClass={cs.AI},
      url={https://arxiv.org/abs/2303.11366}, 
}

@misc{zhao2024expelllmagentsexperiential,
      title={ExpeL: LLM Agents Are Experiential Learners}, 
      author={Andrew Zhao and Daniel Huang and Quentin Xu and Matthieu Lin and Yong-Jin Liu and Gao Huang},
      year={2024},
      eprint={2308.10144},
      archivePrefix={arXiv},
      primaryClass={cs.LG},
      url={https://arxiv.org/abs/2308.10144}, 
}

@misc{hu2021loralowrankadaptationlarge,
      title={LoRA: Low-Rank Adaptation of Large Language Models}, 
      author={Edward J. Hu and Yelong Shen and Phillip Wallis and Zeyuan Allen-Zhu and Yuanzhi Li and Shean Wang and Lu Wang and Weizhu Chen},
      year={2021},
      eprint={2106.09685},
      archivePrefix={arXiv},
      primaryClass={cs.CL},
      url={https://arxiv.org/abs/2106.09685}, 
}

@misc{kimiteam2025kimilinearexpressiveefficient,
      title={Kimi Linear: An Expressive, Efficient Attention Architecture}, 
      author={{Kimi Team} and Yu Zhang and Zongyu Lin and Xingcheng Yao and Jiaxi Hu and Fanqing Meng and Chengyin Liu and Xin Men and Songlin Yang and Zhiyuan Li and Wentao Li and Enzhe Lu and Weizhou Liu and Yanru Chen and Weixin Xu and Longhui Yu and Yejie Wang and Yu Fan and Longguang Zhong and Enming Yuan and Dehao Zhang and Yizhi Zhang and T. Y. Liu and Haiming Wang and Shengjun Fang and Weiran He and Shaowei Liu and Yiwei Li and Jianlin Su and Jiezhong Qiu and Bo Pang and Junjie Yan and Zhejun Jiang and Weixiao Huang and Bohong Yin and Jiacheng You and Chu Wei and Zhengtao Wang and Chao Hong and Yutian Chen and Guanduo Chen and Yucheng Wang and Huabin Zheng and Feng Wang and Yibo Liu and Mengnan Dong and Zheng Zhang and Siyuan Pan and Wenhao Wu and Yuhao Wu and Longyu Guan and Jiawen Tao and Guohong Fu and Xinran Xu and Yuzhi Wang and Guokun Lai and Yuxin Wu and Xinyu Zhou and Zhilin Yang and Yulun Du},
      year={2025},
      eprint={2510.26692},
      archivePrefix={arXiv},
      primaryClass={cs.CL},
      url={https://arxiv.org/abs/2510.26692}, 
}

@misc{lee2026mintevalevaluatingmemorymultitarget,
      title={MINTEval: Evaluating Memory under Multi-Target Interference in Long-Horizon Agent Systems}, 
      author={Hyunji Lee and Justin Chih-Yao Chen and Joykirat Singh and Zaid Khan and Elias Stengel-Eskin and Mohit Bansal},
      year={2026},
      eprint={2605.18565},
      archivePrefix={arXiv},
      primaryClass={cs.CL},
      url={https://arxiv.org/abs/2605.18565}, 
}

@misc{kočiský2017narrativeqareadingcomprehensionchallenge,
      title={The NarrativeQA Reading Comprehension Challenge}, 
      author={Tomáš Kočiský and Jonathan Schwarz and Phil Blunsom and Chris Dyer and Karl Moritz Hermann and Gábor Melis and Edward Grefenstette},
      year={2017},
      eprint={1712.07040},
      archivePrefix={arXiv},
      primaryClass={cs.CL},
      url={https://arxiv.org/abs/1712.07040}, 
}

@misc{dasigi2021datasetinformationseekingquestionsanswers,
      title={A Dataset of Information-Seeking Questions and Answers Anchored in Research Papers}, 
      author={Pradeep Dasigi and Kyle Lo and Iz Beltagy and Arman Cohan and Noah A. Smith and Matt Gardner},
      year={2021},
      eprint={2105.03011},
      archivePrefix={arXiv},
      primaryClass={cs.CL},
      url={https://arxiv.org/abs/2105.03011}, 
}

@misc{sprague2024musrtestinglimitschainofthought,
      title={MuSR: Testing the Limits of Chain-of-thought with Multistep Soft Reasoning}, 
      author={Zayne Sprague and Xi Ye and Kaj Bostrom and Swarat Chaudhuri and Greg Durrett},
      year={2024},
      eprint={2310.16049},
      archivePrefix={arXiv},
      primaryClass={cs.CL},
      url={https://arxiv.org/abs/2310.16049}, 
}

@misc{yang2018hotpotqadatasetdiverseexplainable,
      title={HotpotQA: A Dataset for Diverse, Explainable Multi-hop Question Answering}, 
      author={Zhilin Yang and Peng Qi and Saizheng Zhang and Yoshua Bengio and William W. Cohen and Ruslan Salakhutdinov and Christopher D. Manning},
      year={2018},
      eprint={1809.09600},
      archivePrefix={arXiv},
      primaryClass={cs.CL},
      url={https://arxiv.org/abs/1809.09600}, 
}

@misc{dua2019dropreadingcomprehensionbenchmark,
      title={DROP: A Reading Comprehension Benchmark Requiring Discrete Reasoning Over Paragraphs}, 
      author={Dheeru Dua and Yizhong Wang and Pradeep Dasigi and Gabriel Stanovsky and Sameer Singh and Matt Gardner},
      year={2019},
      eprint={1903.00161},
      archivePrefix={arXiv},
      primaryClass={cs.CL},
      url={https://arxiv.org/abs/1903.00161}, 
}

@misc{chevalierboisvert2019babyaiplatformstudysample,
      title={BabyAI: A Platform to Study the Sample Efficiency of Grounded Language Learning}, 
      author={Maxime Chevalier-Boisvert and Dzmitry Bahdanau and Salem Lahlou and Lucas Willems and Chitwan Saharia and Thien Huu Nguyen and Yoshua Bengio},
      year={2019},
      eprint={1810.08272},
      archivePrefix={arXiv},
      primaryClass={cs.AI},
      url={https://arxiv.org/abs/1810.08272}, 
}

@misc{côté2019textworldlearningenvironmenttextbased,
      title={TextWorld: A Learning Environment for Text-based Games}, 
      author={Marc-Alexandre Côté and Ákos Kádár and Xingdi Yuan and Ben Kybartas and Tavian Barnes and Emery Fine and James Moore and Ruo Yu Tao and Matthew Hausknecht and Layla El Asri and Mahmoud Adada and Wendy Tay and Adam Trischler},
      year={2019},
      eprint={1806.11532},
      archivePrefix={arXiv},
      primaryClass={cs.LG},
      url={https://arxiv.org/abs/1806.11532}, 
}

@misc{mu2024learningcompresspromptsgist,
      title={Learning to Compress Prompts with Gist Tokens}, 
      author={Jesse Mu and Xiang Lisa Li and Noah Goodman},
      year={2024},
      eprint={2304.08467},
      archivePrefix={arXiv},
      primaryClass={cs.CL},
      url={https://arxiv.org/abs/2304.08467}, 
}

@misc{ge2024incontextautoencodercontextcompression,
      title={In-context Autoencoder for Context Compression in a Large Language Model}, 
      author={Tao Ge and Jing Hu and Lei Wang and Xun Wang and Si-Qing Chen and Furu Wei},
      year={2024},
      eprint={2307.06945},
      archivePrefix={arXiv},
      primaryClass={cs.CL},
      url={https://arxiv.org/abs/2307.06945}, 
}

@misc{li2024500xcompressorgeneralizedpromptcompression,
      title={500xCompressor: Generalized Prompt Compression for Large Language Models}, 
      author={Zongqian Li and Yixuan Su and Nigel Collier},
      year={2024},
      eprint={2408.03094},
      archivePrefix={arXiv},
      primaryClass={cs.CL},
      url={https://arxiv.org/abs/2408.03094}, 
}

@misc{ren2025sambasimplehybridstate,
      title={Samba: Simple Hybrid State Space Models for Efficient Unlimited Context Language Modeling}, 
      author={Liliang Ren and Yang Liu and Yadong Lu and Yelong Shen and Chen Liang and Weizhu Chen},
      year={2025},
      eprint={2406.07522},
      archivePrefix={arXiv},
      primaryClass={cs.CL},
      url={https://arxiv.org/abs/2406.07522}, 
}

@misc{dong2024hymbahybridheadarchitecturesmall,
      title={Hymba: A Hybrid-head Architecture for Small Language Models}, 
      author={Xin Dong and Yonggan Fu and Shizhe Diao and Wonmin Byeon and Zijia Chen and Ameya Sunil Mahabaleshwarkar and Shih-Yang Liu and Matthijs Van Keirsbilck and Min-Hung Chen and Yoshi Suhara and Yingyan Lin and Jan Kautz and Pavlo Molchanov},
      year={2024},
      eprint={2411.13676},
      archivePrefix={arXiv},
      primaryClass={cs.CL},
      url={https://arxiv.org/abs/2411.13676}, 
}

@misc{nvidia2025nemotronhfamilyaccurateefficient,
      title={Nemotron-H: A Family of Accurate and Efficient Hybrid Mamba-Transformer Models}, 
      author={NVIDIA and : and Aaron Blakeman and Aarti Basant and Abhinav Khattar and Adithya Renduchintala and Akhiad Bercovich and Aleksander Ficek and Alexis Bjorlin and Ali Taghibakhshi and Amala Sanjay Deshmukh and Ameya Sunil Mahabaleshwarkar and Andrew Tao and Anna Shors and Ashwath Aithal and Ashwin Poojary and Ayush Dattagupta and Balaram Buddharaju and Bobby Chen and Boris Ginsburg and Boxin Wang and Brandon Norick and Brian Butterfield and Bryan Catanzaro and Carlo del Mundo and Chengyu Dong and Christine Harvey and Christopher Parisien and Dan Su and Daniel Korzekwa and Danny Yin and Daria Gitman and David Mosallanezhad and Deepak Narayanan and Denys Fridman and Dima Rekesh and Ding Ma and Dmytro Pykhtar and Dong Ahn and Duncan Riach and Dusan Stosic and Eileen Long and Elad Segal and Ellie Evans and Eric Chung and Erick Galinkin and Evelina Bakhturina and Ewa Dobrowolska and Fei Jia and Fuxiao Liu and Gargi Prasad and Gerald Shen and Guilin Liu and Guo Chen and Haifeng Qian and Helen Ngo and Hongbin Liu and Hui Li and Igor Gitman and Ilia Karmanov and Ivan Moshkov and Izik Golan and Jan Kautz and Jane Polak Scowcroft and Jared Casper and Jarno Seppanen and Jason Lu and Jason Sewall and Jiaqi Zeng and Jiaxuan You and Jimmy Zhang and Jing Zhang and Jining Huang and Jinze Xue and Jocelyn Huang and Joey Conway and John Kamalu and Jon Barker and Jonathan Cohen and Joseph Jennings and Jupinder Parmar and Karan Sapra and Kari Briski and Kateryna Chumachenko and Katherine Luna and Keshav Santhanam and Kezhi Kong and Kirthi Sivamani and Krzysztof Pawelec and Kumar Anik and Kunlun Li and Lawrence McAfee and Leon Derczynski and Lindsey Pavao and Luis Vega and Lukas Voegtle and Maciej Bala and Maer Rodrigues de Melo and Makesh Narsimhan Sreedhar and Marcin Chochowski and Markus Kliegl and Marta Stepniewska-Dziubinska and Matthieu Le and Matvei Novikov and Mehrzad Samadi and Michael Andersch and Michael Evans and Miguel Martinez and Mike Chrzanowski and Mike Ranzinger and Mikolaj Blaz and Misha Smelyanskiy and Mohamed Fawzy and Mohammad Shoeybi and Mostofa Patwary and Nayeon Lee and Nima Tajbakhsh and Ning Xu and Oleg Rybakov and Oleksii Kuchaiev and Olivier Delalleau and Osvald Nitski and Parth Chadha and Pasha Shamis and Paulius Micikevicius and Pavlo Molchanov and Peter Dykas and Philipp Fischer and Pierre-Yves Aquilanti and Piotr Bialecki and Prasoon Varshney and Pritam Gundecha and Przemek Tredak and Rabeeh Karimi and Rahul Kandu and Ran El-Yaniv and Raviraj Joshi and Roger Waleffe and Ruoxi Zhang and Sabrina Kavanaugh and Sahil Jain and Samuel Kriman and Sangkug Lym and Sanjeev Satheesh and Saurav Muralidharan and Sean Narenthiran and Selvaraj Anandaraj and Seonmyeong Bak and Sergey Kashirsky and Seungju Han and Shantanu Acharya and Shaona Ghosh and Sharath Turuvekere Sreenivas and Sharon Clay and Shelby Thomas and Shrimai Prabhumoye and Shubham Pachori and Shubham Toshniwal and Shyamala Prayaga and Siddhartha Jain and Sirshak Das and Slawek Kierat and Somshubra Majumdar and Song Han and Soumye Singhal and Sriharsha Niverty and Stefania Alborghetti and Suseella Panguluri and Swetha Bhendigeri and Syeda Nahida Akter and Szymon Migacz and Tal Shiri and Terry Kong and Timo Roman and Tomer Ronen and Trisha Saar and Tugrul Konuk and Tuomas Rintamaki and Tyler Poon and Ushnish De and Vahid Noroozi and Varun Singh and Vijay Korthikanti and Vitaly Kurin and Wasi Uddin Ahmad and Wei Du and Wei Ping and Wenliang Dai and Wonmin Byeon and Xiaowei Ren and Yao Xu and Yejin Choi and Yian Zhang and Ying Lin and Yoshi Suhara and Zhiding Yu and Zhiqi Li and Zhiyu Li and Zhongbo Zhu and Zhuolin Yang and Zijia Chen},
      year={2025},
      eprint={2504.03624},
      archivePrefix={arXiv},
      primaryClass={cs.CL},
      url={https://arxiv.org/abs/2504.03624}, 
}

@misc{michalak2025attentionneedretrieval,
      title={Some Attention is All You Need for Retrieval}, 
      author={Felix Michalak and Steven Abreu},
      year={2025},
      eprint={2510.19861},
      archivePrefix={arXiv},
      primaryClass={cs.LG},
      url={https://arxiv.org/abs/2510.19861}, 
}

@misc{borobia2026componentablationefficienthybrid,
      title={Component Ablation for Efficient Hybrid Language Model Architectures: Performance, Resilience, and Compression Implications}, 
      author={Hector Borobia and Elies Seguí-Mas and Guillermina Tormo-Carbó},
      year={2026},
      eprint={2603.22473},
      archivePrefix={arXiv},
      primaryClass={cs.CL},
      url={https://arxiv.org/abs/2603.22473}, 
}

@misc{bick2026retrievalawaredistillationtransformerssmhybrids,
      title={Retrieval-Aware Distillation for Transformer-SSM Hybrids}, 
      author={Aviv Bick and Eric P. Xing and Albert Gu},
      year={2026},
      eprint={2602.11374},
      archivePrefix={arXiv},
      primaryClass={cs.LG},
      url={https://arxiv.org/abs/2602.11374}, 
}

@misc{lee2025understandingenhancingmambatransformerhybrids,
      title={Understanding and Enhancing Mamba-Transformer Hybrids for Memory Recall and Language Modeling}, 
      author={Hyunji Lee and Wenhao Yu and Hongming Zhang and Kaixin Ma and Jiyeon Kim and Dong Yu and Minjoon Seo},
      year={2025},
      eprint={2510.26912},
      archivePrefix={arXiv},
      primaryClass={cs.CL},
      url={https://arxiv.org/abs/2510.26912}, 
}

@misc{zhang2026metismemoryfoundationmodel,
      title={Metis: Memory Foundation Model}, 
      author={Zeyu Zhang and Ziliang Guo and Yihang Sun and Xichong Zhang and Xixuan Hao and Zehao Lin and Yang Zhang and Xiaoyan Zhao and Tong Shen and Bo Tang and Zhi-Qin John Xu and Junchi Yan and Haofen Wang and Xu Chen and Feiyu Xiong and Zhiyu Li and Tat-Seng Chua},
      year={2026},
      eprint={2607.26760},
      archivePrefix={arXiv},
      primaryClass={cs.CL},
      url={https://arxiv.org/abs/2607.26760}, 
}

@misc{dao2024transformersssmsgeneralizedmodels,
      title={Transformers are SSMs: Generalized Models and Efficient Algorithms Through Structured State Space Duality}, 
      author={Tri Dao and Albert Gu},
      year={2024},
      eprint={2405.21060},
      archivePrefix={arXiv},
      primaryClass={cs.LG},
      url={https://arxiv.org/abs/2405.21060}, 
}

@misc{yang2025gateddeltanetworksimproving,
      title={Gated Delta Networks: Improving Mamba2 with Delta Rule}, 
      author={Songlin Yang and Jan Kautz and Ali Hatamizadeh},
      year={2025},
      eprint={2412.06464},
      archivePrefix={arXiv},
      primaryClass={cs.CL},
      url={https://arxiv.org/abs/2412.06464}, 
}
\bibliographystyle{iclr2027_conference}

\appendix
\newpage

\section{Details on Role analysis of hybrid LMs}
\label{app:role}

\subsection{Verbatim vs. Non-Verbatim Questions}
We first distinguish questions based on whether their answers can be directly recovered from the source context. 
We define a question as \emph{verbatim} when its normalized gold answer appears in the source document and \emph{non-verbatim} otherwise.
Approximately 34.6\% of the test examples (about 2.8k examples) are verbatim and to match the number of instances for the two categories, we randomly sample the same number of examples from the remaining non-verbatim questions.

\subsection{Single- vs. Multi-Evidence Questions}
We next examine whether the relative strengths of the two memory pathways change with the amount of evidence required to answer a question.
Using the evidence annotations provided by the datasets, we divide questions into those that rely on a single piece of evidence and those that require multiple pieces of evidence.
Among the test examples, approximately 30.2\% (about 2.5k) are single-evidence and 39.7\% (about 3.3k) are multi-evidence, and the rest does not provide explicit evidence annotations. 
To balance the two categories, we randomly sample the same number of multi-evidence questions as single-evidence ones, resulting in around 2.5k examples each.

\section{Details on Experimental Setup}
\label{app:exp_setup}

\subsection{Question Answering}
We evaluate each dataset with its standard metric after text normalization: token-level F1 for NarrativeQA and Qasper, and exact match (EM) for DROP, HotpotQA, and MINTEval.
For MuSR, we provide all answer options in the input and ask the model to generate one of them, following the same generation format as the other datasets, and compute EM.
For MINTEval, we evaluate on the wiki-revision and git-commit.
For brevity, we refer to these metrics collectively as accuracy in the paper.
We train models for 5 epochs with early stopping.
For evaluation, we subsample large test sets~(DROP, NarrativeQA, HotpotQA, and MINTEval) to 1.5k due to compute cost. 

\subsection{Agentic Setup}
We report success rate, the fraction of games in which the agent completes the goal within the step limit.
For TextWorld, we set the maximum number of steps to 100 and use 1,000 training, 150 validation and 200 test games, generated with non-overlapping random seeds.
For BabyAI, we use the official maximum number of steps for each level with 100 instances for each training level (MiniBossLevel) and test levels (PutNextLocal, GoToObjMaze). 
In the retry experiments (Section~\ref{analysis:agentic}), the agent is allowed up to 7 attempts per game, with its previous trajectories provided in context.
In the demonstration experiments, we run the agent multiple times on each game to build a pool of successful trajectories with varying lengths, and sample five trajectories from this pool as demonstrations.
We train the agents using the gold trajectories for 5 epochs with early stopping.

\section{Additional Analysis}

\subsection{Effect of the auxiliary loss weight $\lambda$}
\label{app_hybrid_model:lambda}

\begin{figure}[t!]
    \begin{minipage}[t]{0.49\linewidth}
    \vspace{0pt}
    \centering
    \includegraphics[width=\linewidth]{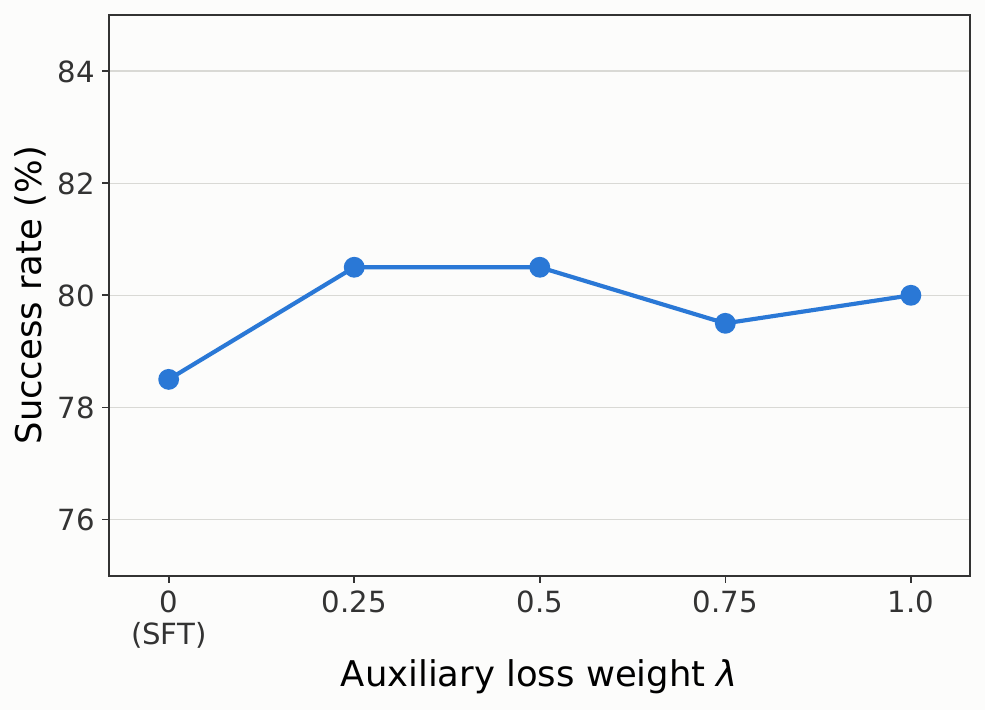}
    \caption{
    Success rate~(y-axis) of Qwen3.5 on TextWorld Quest across different auxiliary loss weights $\lambda$~(x-axis), where $\lambda=0$ denotes standard SFT. The auxiliary loss improves over standard SFT for all $\lambda$.
    }
    \label{fig:lambda_ablation}

    \end{minipage}
\hfill
    \centering
        \begin{minipage}[t]{0.49\linewidth}
    \vspace{0pt}
    \centering
    \includegraphics[width=\linewidth]{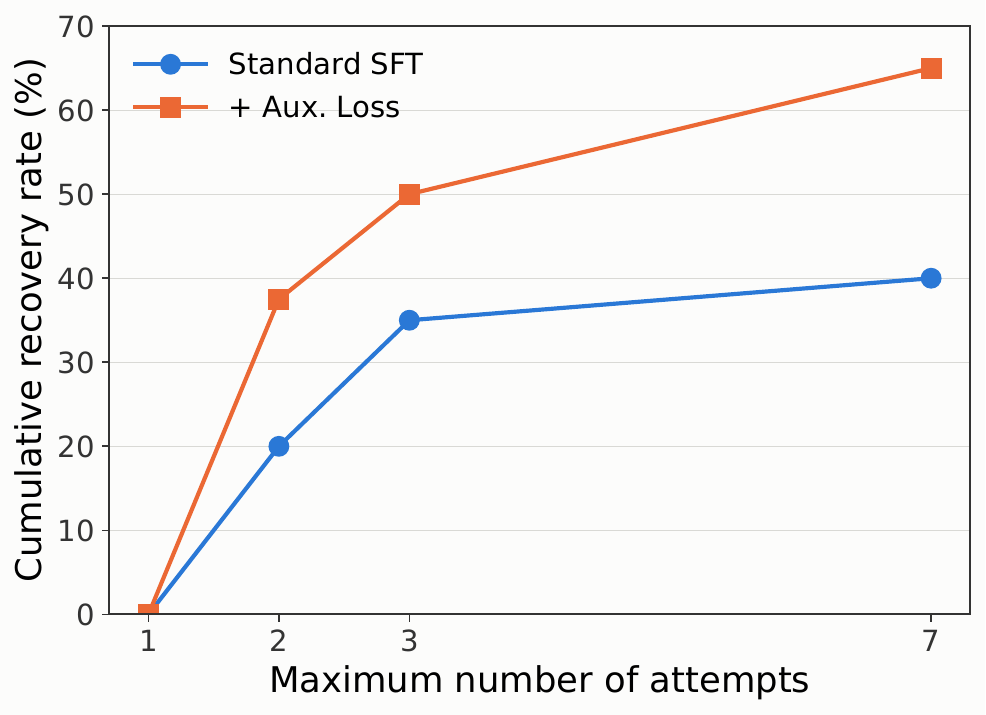}
    \caption{
    Cumulative recovery rate (y-axis) on agentic tasks as the maximum number of attempts increases (x-axis), where previous failed trajectories are provided in context. The auxiliary loss continues to improve as attempts accumulate, whereas standard SFT plateaus.
    }
    \label{fig:agentic_learning_from_failure}
    \end{minipage}
    
\end{figure}

Figure~\ref{fig:lambda_ablation} shows the success rate of Qwen3.5 on TextWorld Quest as we vary the weight of the auxiliary loss $\lambda$.
Adding the auxiliary loss consistently improves over standard SFT ($\lambda=0$, 78.5\%) for all values of $\lambda$.
We use $\lambda=0.25$, which achieves the best performance, in all experiments unless otherwise specified.

\subsection{Agentic Tasks}

We examine whether agents can learn from their own failures: when the agent fails to complete a game, it retries the same game up to a maximum number of attempts, with its previous trajectories provided in context.
When evaluating over TextWorld, as shown in Figure~\ref{fig:agentic_learning_from_failure}, the auxiliary loss improves the cumulative recovery rate by 19.2\% on average. 
Notably, the standard SFT model tends to converge as the number of attempts increases, whereas the model with the auxiliary loss continues to improve, suggesting that it learns better from past failure trajectories.

\subsection{Generalizes to Attention-Based Models}
\label{app:attention_based}

To examine whether our findings generalize to attention-based models, which do not have multiple built-in memory pathways, we provide them with two commonly used forms of memory: textual~\citep{packer2024memgptllmsoperatingsystems, yu2026agenticmemorylearningunified, wang2025memalphalearningmemoryconstruction} and latent~\citep{mu2024learningcompresspromptsgist, li2024500xcompressorgeneralizedpromptcompression} memory. We conduct the experiment with \texttt{Qwen3-4B-Instruct} model.

\subsubsection{Textual \& Latent Memory.}
\label{app:attn_based_textual_latent}
For \textbf{textual memory}, we summarize a given document and train the model to generate the answer conditioned on the resulting summary and the question. 
For \textbf{latent memory}, following prior work~\citep{mu2024learningcompresspromptsgist, ge2024incontextautoencodercontextcompression}, we represent each document as a layer-wise KV prefix with $K$ memory slots.
Specifically, we append $K$ learnable memory tokens to the end of the document and run the resulting sequence through the model.
Because these memory tokens follow the document, they can attend to and aggregate information from the preceding tokens. 
We retain the key-value representations at these $K$ memory tokens positions from every transformer layer; together, these representations form the latent memory. 
Training proceeds in two stages: in Stage 1, we train the model to encode document information into these latent memory slots using reasoning supervision.
In Stage 2, we provide the latent memory as a prefix cache to the model, which is trained to generate the answer conditioned on the latent memory and the question. 
The answer-generation loss is backpropagated through the entire computation, including the forward pass used to construct the latent memory.

\vspace{10em}
\paragraph{Textual and Latent Memory Play Complementary Roles.}

\begin{wrapfigure}{r}{0.45\linewidth}
    \centering
    \vspace{-13pt}
    \includegraphics[width=\linewidth]{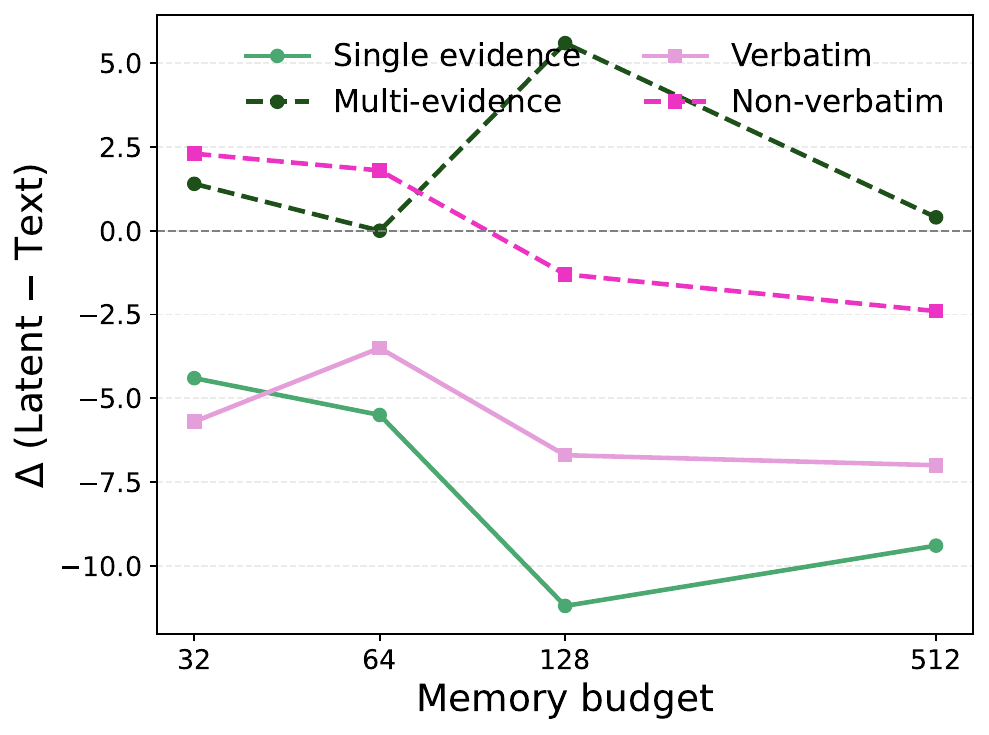}
    \caption{
    Relative performance of latent versus textual memory (y-axis) under matched memory budgets (x-axis). Textual memory performs better on verbatim and single-evidence questions, while latent memory is more competitive on non-verbatim and multi-evidence questions.
    }
    \label{fig:diff_info}
\end{wrapfigure}

We compare textual and latent memory under a range of matched memory budgets along two axes: whether the answer appears word-for-word in the context (verbatim vs. non-verbatim) and whether answering requires one or multiple pieces of evidence (single- vs. multi-evidence). See Appendix~\ref{app:role} for details.
As shown in Figure~\ref{fig:diff_info}, textual memory shows a larger advantage when the required information is directly recoverable (verbatim) or localized in a single piece of evidence (single-evidence), whereas latent memory becomes more competitive when the answer must be inferred (non-verbatim) or requires combining multiple pieces of evidence (multi-evidence).
This suggests that textual memory is especially effective when prediction relies on precise information explicitly retained in memory, whereas latent memory is relatively more effective when information must be transformed or aggregated.

\subsubsection{Standard SFT Favors Textual Memory over Latent Memory}
\label{attn_based:explicit_coordination}

\begin{observationbox}
\textbf{Key Observation:} Naively training hybrid memory with standard SFT fails to coordinate its components effectively, instead favoring textual memory.
\end{observationbox}

\paragraph{Naive Hybrid Memory.}
For \textit{naive hybrid memory}, we follow the same Stage 1 as for latent memory, training the model with reasoning supervision to encode document information into the $K$ latent memory slots.
In Stage 2, we provide both the latent memory and textual memory to the model and train it with the same question-answering objective.
Thus, both memory representations are \textit{jointly} available during training, without any additional supervision that explicitly encourages coordination between them.

\vspace{10pt}
\paragraph{Analysis.}
Figure~\ref{fig:hybrid_memory_all} shows question answering performance with different types of memory, using a textual memory budget of 8k tokens~\citep{wang2025memalphalearningmemoryconstruction} and a latent memory budget of 128 slots~\citep{ge2024incontextautoencodercontextcompression}.
The \textit{Oracle}, which selects the correct prediction whenever either single-memory model succeeds, measures the complementary information available across the two memories.
It outperforms the single-memory models~(\textit{Text} and \textit{Latent}) by 10.1\% on average, indicating that the two memories capture complementary information.
However, the model trained with both memories under standard SFT (\textit{SFT}) remains 7.9\% below the \textit{Oracle} and even falls below the \textit{Text} (24.1\% vs.\ 25.9\%), suggesting that naively combining the two memories does not ensure better coordination or improved performance, and can instead lead them to interfere with each other. 
These results show that \textbf{the presence of complementary information across memories does not by itself guarantee effective coordination under joint training}.

\subsubsection{Adding an Auxiliary Loss to Encourage Latent Memory Use Improves Performance}
\label{attn_based:learn_coordination}

\begin{observationbox}
\textbf{Key Observation:} 
Adding an auxiliary loss that encourages the model to use a designated memory pathway improves memory coordination and overall performance. The gains are largest when pathway selection follows the \textit{verbatim} criterion identified in Appendix~\ref{app:attn_based_textual_latent}.
\end{observationbox}

The previous section shows that naive joint training leads to imbalanced use of the two memory pathways, with the model relying predominantly on textual memory and underutilizing information captured by latent memory. 
We therefore examine whether an auxiliary objective that explicitly encourages the model to make use of the underutilized latent memory improves memory coordination and overall performance.

\paragraph{Auxiliary Loss for Memory Coordination.}

To examine whether encouraging the use of the underutilized latent memory improves memory coordination and, in turn, improves overall performance, we add an auxiliary loss that assigns each memory an explicit role based on its strengths~(Appendix~\ref{app:attn_based_textual_latent}).

The auxiliary loss encourages the model to use a designated memory pathway by making the other pathway uninformative for the target. 
For each example $i$, the standard hybrid-memory objective is:
\[
\ell(T_i,L_i)
=
-\log p_\theta(y_i \mid q_i, T_i, L_i),
\]
where $q_i$ and $y_i$ denote the query and target, and $T_i$ and $L_i$ denote the textual and latent memories constructed from the source context of example $i$.

We augment the objective with an auxiliary term~($\mathcal{L}^{\mathrm{aux}}_i$):
\begin{equation}
\mathcal{L}_i
=
\ell(T_i,L_i)
+
\lambda \mathcal{L}^{\mathrm{aux}}_i,
\qquad
\mathcal{L}^{\mathrm{aux}}_i
=
\begin{cases}
\ell(T_i,\operatorname{sg}[L_j]), & \mathrm{role}(i)=\mathrm{text},\\
\ell(T_j,L_i), & \mathrm{role}(i)=\mathrm{latent}.
\end{cases}
\end{equation}

Here, $\lambda$ controls the weight of the auxiliary loss, and $j$ denotes a sampled irrelevant example\footnote{To reduce the unrelated factor impact, we sample $j$ from the same source dataset with a similar document length.}.
For a \textit{text-role} example~($\mathrm{role}(i)=\mathrm{text}$), we keep the correct textual memory $T_i$ and replace the latent memory with $L_j$, so that the model must rely on $T_i$ to predict $y_i$; we stop gradients~($\mathrm{sg}[\cdot]$) through $L_j$ so that this auxiliary term does not update the latent-memory writer.
For a \textit{latent-role} example~($\mathrm{role}(i)=\mathrm{latent}$), we instead keep the correct latent memory $L_i$ and replace the textual memory with $T_j$, encouraging the model to preserve and use the target-relevant information through the latent pathway.
We use the same $\lambda$ to 0.25 for all experiments.

Since the optimal role for each example is not known a priori, we experiment over four criteria for assigning $\mathrm{role}(i)$.
\textsc{Outcome} assigns roles based on which single-memory model answers the example correctly: $\mathrm{role}(i)=\mathrm{latent}$ if the latent-memory model answers example $i$ correctly, and $\mathrm{role}(i)=\mathrm{text}$ otherwise.
\textsc{Random} assigns each example to the text or latent role uniformly at random.
\textsc{Verbatim} follows the distinction identified in Appendix~\ref{app:attn_based_textual_latent}, assigning verbatim examples to the text role and non-verbatim examples to the latent role: $\mathrm{role}(i)=\mathrm{text}$ if $i$ = verbatim else $\mathrm{role}(i)=\mathrm{latent}$. 
\textsc{Reverse} uses the opposite assignment, assigning verbatim examples to the latent role and non-verbatim examples to the text role: $\mathrm{role}(i)=\mathrm{latent}$ if $i$ = verbatim else $\mathrm{role}(i)=\mathrm{text}$.

\begin{table}[t!]
    \centering
    \small
    \setlength{\tabcolsep}{4pt}
    \caption{
    Accuracy and memory use across auxiliary-training variants.
    $\Delta$Latent and $\Delta$Text denote the accuracy drop when the latent or textual memory, respectively, is replaced with one constructed from an unrelated context; higher values indicate greater use of that memory.
    \textit{Role-conditioned} indicates whether auxiliary supervision uses per-example role assignment.
    }
    \label{table:role_assignment}
    \begin{tabular}{llGcc}
        \toprule
        Role-conditioned & Variant & Acc. & $\Delta$Latent & $\Delta$Text \\
        \midrule
        - & \textsc{Naive} & 24.1 & -2.4 & +11.4 \\
        \midrule
        \multirow{4}{*}{Yes} & \textsc{Outcome}  & 26.1 & +0.8 & +11.5 \\
                             & \textsc{Verbatim} & \textbf{28.6} & \textbf{+4.7} & +10.4 \\
                             & \textsc{Reverse}  & 24.5 & +1.1 & +7.2 \\
                             & \textsc{Random}   & 23.9 & -1.0 & +10.5 \\
        \midrule
        \multirow{3}{*}{No}  & \textsc{Latent-aux} & 26.1 & +3.4 & +8.5 \\
                             & \textsc{Text-aux}   & 24.9 & +0.0 & \textbf{+14.5} \\
                             & \textsc{Unrouted}   & 24.4 & -1.2 & +9.1 \\
        \midrule
        - & \textsc{Oracle} & 32.0 & & \\
        \bottomrule
    \end{tabular}
\end{table}

\paragraph{Results.}
We examine whether auxiliary supervision improves coordination between textual and latent memory, and how the criterion used to assign $\mathrm{role}(i)$ affects this coordination.
Table~\ref{table:role_assignment} reports accuracy together with the use of each memory ($\Delta$Latent and $\Delta$Text).
Standard SFT without auxiliary supervision (\textsc{Naive}) largely ignores latent memory: replacing it with an unrelated one even slightly improves accuracy ($\Delta$Latent $=-2.4$).

\textit{Role-conditioned supervision.}
Among the role assignment criteria, \textsc{Verbatim} achieves the best accuracy, improving it from 24.1\% to 28.6\%, and the largest increase in latent-memory use ($-2.4 \rightarrow +4.7$) while largely preserving textual-memory use ($+11.4 \rightarrow +10.4$).
\textsc{Outcome} also improves accuracy (26.1\%) but yields a smaller increase in latent-memory use; we hypothesize that per-example correctness is influenced by factors beyond the consistent strengths of each memory, making it a noisier signal for role assignment.
\textsc{Reverse} increases latent-memory use only modestly while reducing textual-memory use, and \textsc{Random} does not improve accuracy, showing that the assignment criterion is critical for coordinating the two memories.

\textit{Supervision without role assignment.}
We also apply auxiliary supervision without per-example role assignment: \textsc{Latent-aux} and \textsc{Text-aux} assign all examples to the latent or text role, respectively, and \textsc{Unrouted} applies both auxiliary terms to every example.
\textsc{Latent-aux} increases latent-memory use and improves accuracy to 26.1\%, but at the cost of textual-memory use, whereas \textsc{Text-aux} mainly increases textual-memory use with little gain in accuracy, and \textsc{Unrouted} improves neither memory use nor accuracy.
Together, these results suggest that increasing the use of a single memory can help, but the largest gains come from assigning each example to the memory suited to it, according to the complementary strengths of the two memories.

\end{document}